\documentclass[11pt,a4paper]{article}
\usepackage[hyperref]{emnlp-ijcnlp-2019}
\usepackage{times}
\usepackage{latexsym}

\usepackage{graphicx}
\usepackage{amsmath}
\usepackage{algorithm}
\usepackage{algorithmic}
\usepackage{enumerate}
\usepackage{mathrsfs}
\usepackage{multirow}
\usepackage{framed}
\usepackage{color}

\usepackage{url}

\aclfinalcopy % Uncomment this line for the final submission

\title{Contrastive Attention Mechanism for Abstractive Sentence Summarization}

\author{
	{Xiangyu Duan\textsuperscript{1,2}\thanks{  $\quad $ Equal contribution. } , Hongfei Yu\textsuperscript{2$\ast$}, Mingming Yin\textsuperscript{2}, Min Zhang\textsuperscript{1,2}, Weihua Luo\textsuperscript{3}, Yue Zhang\textsuperscript{4} }
	\vspace{1.6mm}\\
	\fontsize{12}{10}\selectfont\itshape
	\,\textsuperscript{\rm 1}  Institute of Artificial Intelligence, Soochow University, Suzhou, China  \\
	    \fontsize{12}{10}\selectfont\itshape  \textsuperscript{\rm 2} School of Computer Science and Technology, Soochow University, Suzhou, China \\
            \fontsize{12}{10}\selectfont\itshape  \textsuperscript{\rm 3} Alibaba DAMO Academy, Hangzhou, China \\
            \fontsize{12}{10}\selectfont\itshape  \textsuperscript{\rm 4} School of Engineering, Westlake University, China \\
            \fontsize{10}{10}\selectfont xiangyuduan@suda.edu.cn;  \{hfyu,mmyin\}@stu.suda.edu.cn;  minzhang@suda.edu.cn\\
	\fontsize{10}{10}\selectfont weihua.luowh@alibaba-inc.com; yue.zhang@wias.org.cn\\	}

\date{}

\begin{document}
\maketitle
\begin{abstract}
We propose a contrastive attention mechanism to extend the sequence-to-sequence framework for abstractive sentence summarization task, which aims to generate a brief summary of a given source sentence. The proposed contrastive attention mechanism accommodates two categories of attention: one is the conventional attention that attends to relevant parts of the source sentence, the other is the opponent attention that attends to irrelevant or less relevant parts of the source sentence. Both attentions are trained in an opposite way so that the contribution from the conventional attention is encouraged and the contribution from the opponent attention is discouraged through a novel softmax and softmin functionality. Experiments on benchmark datasets show that, the proposed contrastive attention mechanism is more focused on the relevant parts for the summary than the conventional attention mechanism, and greatly advances the state-of-the-art performance on the abstractive sentence summarization task. We release the code at \url{https://github.com/travel-go/Abstractive-Text-Summarization}.

\end{abstract}

\section{Introduction}

Abstractive sentence summarization aims at generating concise and informative summaries based on the core meaning of source sentences. Previous endeavors tackle the problem through either rule-based methods ~\cite{Dorr2003Hedge} or statistical models trained on relatively small scale training corpora  ~\cite{Banko2000Headline}. Following its successful applications on machine translation ~\cite{Sutskever2014RNN,Bahdanau2015RNN}, the sequence-to-sequence framework is also applied on the abstractive sentence summarization task using large-scale sentence summary corpora ~\cite{Rush2015Sum,Chopra2016RNN,Nallapati2016Abstractive}, obtaining better performance compared to the traditional methods.

One central component in state-of-the-art sequence to sequence models is the use of attention for building connections between the source sequence and target words, so that a more informed decision can be made for generating a target word by considering the most relevant parts of the source sequence ~\cite{Bahdanau2015RNN,Vaswani2017Attention}. For abstractive sentence summarization, such attention mechanisms can be useful for selecting the most salient words for a short summary, while filtering the negative influence of redundant parts.

We consider improving abstractive summarization quality by enhancing target-to-source attention. In particular, a contrastive mechanism is taken, by encouraging the contribution from the conventional attention that attends to relevant parts of the source sentence, while at the same time penalizing the contribution from an opponent attention that attends to irrelevant or less relevant parts. Contrastive attention was first proposed in computer vision ~\cite{Song2018Mask}, which is used for person re-identification by attending to person and background regions contrastively. To our knowledge, we are the first to use contrastive attention for NLP and deploy it in the sequence-to-sequence framework. 

In particular, we take Transformer ~\cite{Vaswani2017Attention} as the baseline summarization model, and enhance it with a proponent attention module and an opponent attention module. The former acts as the conventional attention mechanism, while the latter can be regarded as a dual module to the former, with similar weight calculation structure, but using a novel softmin function to discourage contributions from irrelevant or less relevant words.

To our knowledge, we are the first to investigate Transformer as a sequence to sequence summarizer. Results on three benchmark datasets show that it gives highly competitive accuracies compared with RNN and CNN alternatives.  When equipped with the proposed contrastive attention mechanism, our Transformer model achieves the best reported results on all data. The visualization of attentions shows that through using the contrastive attention mechanism, our attention is more focused on relevant parts  than the baseline. We release our code at XXX.

\section{Related Work}

Automatic summarization has been investigated in two main paradigms: the extractive method and the abstractive method. The former extracts important pieces of source document and concatenates them sequentially ~\cite{Jing2000Cut,Knight2000Statistics,Neto2002Automatic}, while the latter grasps the core meaning of the source text and re-state it in short text as abstractive summary ~\cite{Banko2000Headline,Rush2015Sum}. In this paper, we focus on abstractive summarization, and especially on abstractive sentence summarization.

Previous work deals with the abstractive sentence summarization task by using either rule based methods ~\cite{Dorr2003Hedge}, or statistical methods utilizing a source-summary parallel corpus to train a machine translation model ~\cite{Banko2000Headline}, or a syntax based transduction model ~\cite{Cohn2008Sentence,Woodsend2010Title}. 

In recent years, sequence-to-sequence neural framework becomes predominant on this task by encoding long source texts and decoding into short summaries together with the attention mechanism. RNN is the most commonly adopted and extensively explored architecture ~\cite{Chopra2016RNN,Nallapati2016Abstractive,Li2017Deep}. A CNN-based architecture is recently employed by Gehring \textit{et al.} \shortcite{Gehring2017Convolutional} using ConvS2S, which applies CNN on both encoder and decoder. Later, Wang \textit{et al.} \shortcite{Wang2018CNN} build upon ConvS2S with topic words embedding and encoding, and train the system with reinforcement learning.

The most related work to our contrastive attention mechanism is in the field of computer vision. Song \textit{et al.} \shortcite{Song2018Mask} first propose the contrastive attention mechanism for person re-identification. In their work, based on a pre-provided person and background segmentation, the two regions are contrastively attended so that they can be easily discriminated. In comparison, we apply the contrastive attention mechanism for sentence level summarization by contrastively attending to relevant parts and irrelevant or less relevant parts. Furthermore, we propose a novel softmax softmin functionality to train the attention mechanism, which is different to Song \textit{et al.} \shortcite{Song2018Mask}, who use mean squared error loss for attention training.

Other explorations with respect to the characteristics of the abstractive summarization task include copying mechanism that copies words from source sequences for composing summaries ~\cite{Gu2016Incorporating,Gulcehre2016Pointing,Song2018Structure}, the selection mechanism that elaborately selects important parts of source sentences ~\cite{Zhou2017Selective,Lin2018Global}, the distraction mechanism that avoids repeated attention on the same area ~\cite{Chen2016Distraction}, and the sequence level training that avoids exposure bias in teacher forcing methods ~\cite{Ayana2016NHG,Li2018Actor,Edunov2018Classical}. Such methods are built on conventional attention, and are orthogonal to our proposed contrastive attention mechanism. 

\section{Approach}

We use two categories of attention for summary generation. One is the conventional attention that attends to relevant parts of source sentence, the other is the opponent attention that contrarily attends to irrelevant or less relevant parts. Both categories of attention output probability distributions over summary words, which are jointly optimized by encouraging the contribution from the conventional attention and discouraging the contribution from the opponent attention.

\begin{figure}[htbp]
\flushleft
\includegraphics[height=7cm,width=8cm]{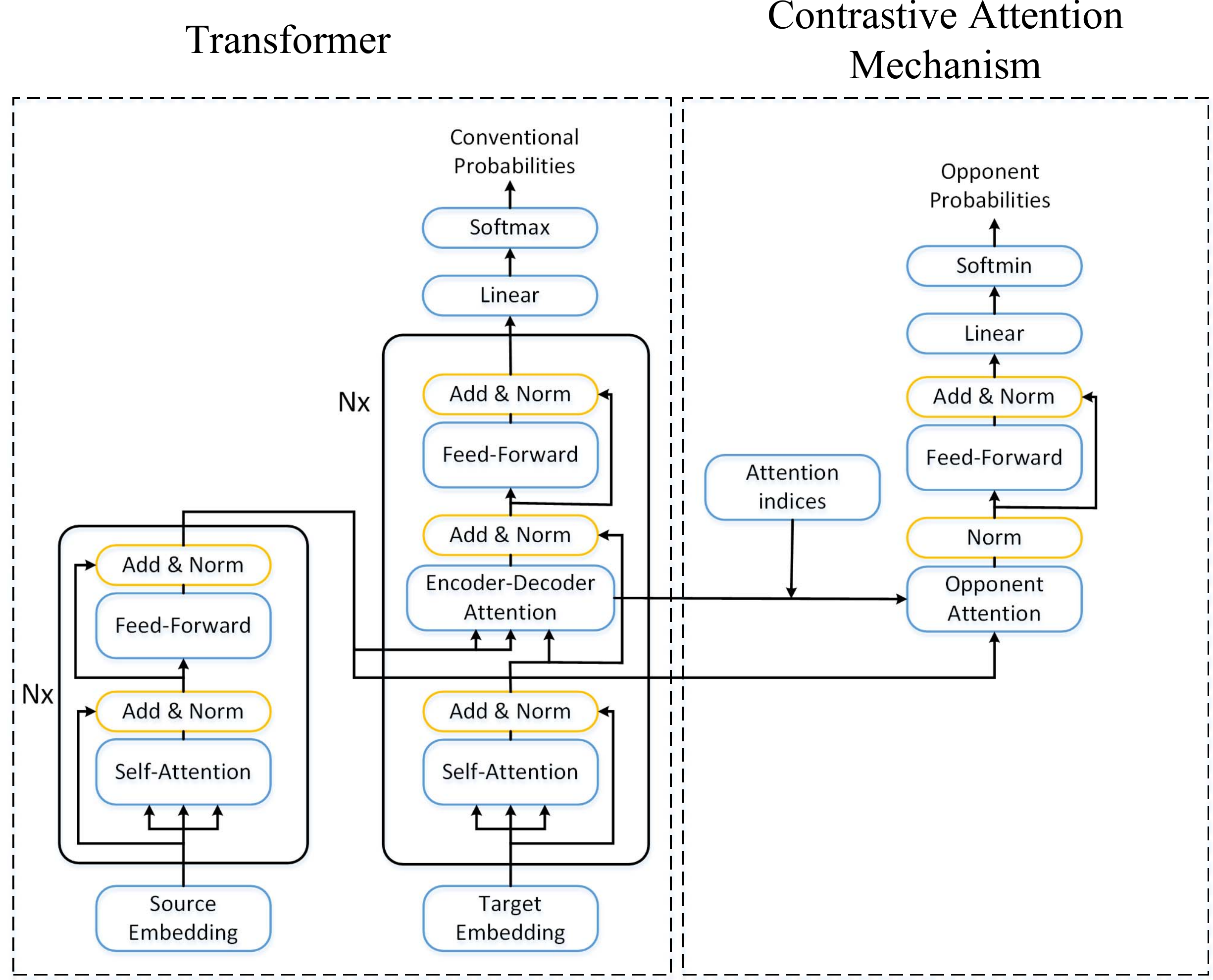}
\caption{Overall networks. The left part is the original Transformer. The right part that takes the opponent attention as bottom layer fulfils the contrastive attention mechanism.} 
\label{fig:overall}
\end{figure}

Figure \ref{fig:overall} illustrates the overall networks. We use Transformer architecture as our basis, upon which we build the contrastive attention mechanism. The left part is the original Transformer. We derive the opponent attention from the conventional attention which is the encoder-decoder attention of the original Transformer, and stack several layers on top of the opponent attention as shown in the right part of Figure \ref{fig:overall}. Both parts contribute to the summary generation by producing probability distributions over the target vocabulary, respectively. The left part outputs {\bf the conventional probability} based on the conventional attention as the original Transformer does, while the right part outputs {\bf the opponent probability} based on the opponent attention. The two probabilities in Figure \ref{fig:overall}  are jointly optimized in a novel way as explained in Section 3.3.

\subsection{Transformer for Abstractive Sentence Summarization}

Transformer is an attention network based sequence-to-sequence architecture ~\cite{Vaswani2017Attention}, which encodes the source text into hidden vectors and decodes into the target text based on the source side information and the target generation history. In comparison to the RNN based architecture and the CNN based architecture, both the encoder and the decoder of Transformer adopt attention as main function.

Let $X$ and $Y$ denote the source sentence and its summary, respectively. Transformer is trained to maximize the probability of $Y$ given $X$: $\prod_i{P_c(y_i|y_1^{i-1},X)}$, where $P_c(y_i|y_1^{i-1},X)$ is the conventional probability of the current summary word $y_i$ given the source sentence and the summary generation history. $P_c$ is computed based on the attention mechanism and the stacked deep layers as shown in the left part of Figure \ref{fig:overall}.

\vspace{6 pt}
\noindent  \textbf{Attention Mechanism}

\noindent Scaled dot-product attention is applied in Transformer:

\begin{equation}
{\rm attention}(Q,K,V)={\rm softmax}(\frac{QK^{\rm T}}{\sqrt{d_k}}) V        \label{equ:general_att}
\end{equation}

\noindent where $Q,K,V$ denotes query vector, key vectors, and value vectors, respectively. $d_k$ denotes the dimension of one vector of $K$. Softmax function outputs the attention weights distributed over $V$. ${\rm attention}(Q,K,V$) is a vector of weighted sum of elements of $V$, and represents current context information. 

We focus on the encoder-decoder attention, which builds the connection between source and target by informing the decoder which area of the source text should be attended to.  Specifically, in the encoder-decoder attention, $Q$ is the single vector coming from the current position of the decoder, $K$ and $V$ are the same sequence of vectors that are the outcomes of the encoder at all source positions. Softmax function distributes the {\bf attention weights} over the source positions.

The attentions in Transformer adopts the multi-head implementation, in which each head computes attention as Equation (\ref{equ:general_att}) but with smaller $Q,K,V$ whose dimension is $1/h$ times of their original dimension respectively. The attentions from $h$ heads are concatenated together and linearly projected to compose the final attention. In this way, multi-head attention provides a multi-view of attention behavior beneficial for the final performance. 

\vspace{6 pt}
\noindent  \textbf{Deep Layers}

\noindent The ``N$\times$" plates in Figure \ref{fig:overall} stands for the stacked N identical layers. On the source side, each layer of the stacked N layers contains two sublayers: the self-attention mechanism, and the fully connected feed-forward network. Each sublayer employs residual connection that adds input to outcome of sublayer, then layer normalization is employed on the outcome of the residual connection.

On the target summary side, each layer contains an additional sublayer of the encoder-decoder attention between the self-attention sublayer and the feed-forward sublayer. At the top of the decoder, the softmax layer is applied to convert the decoder output to summary word generation probabilities.  

\subsection{Contrastive Attention Mechanism}

\subsubsection{Opponent Attention} \label{sec:opponent_att}

\begin{figure}[htbp]
\flushleft
\includegraphics[height=5cm,width=8cm]{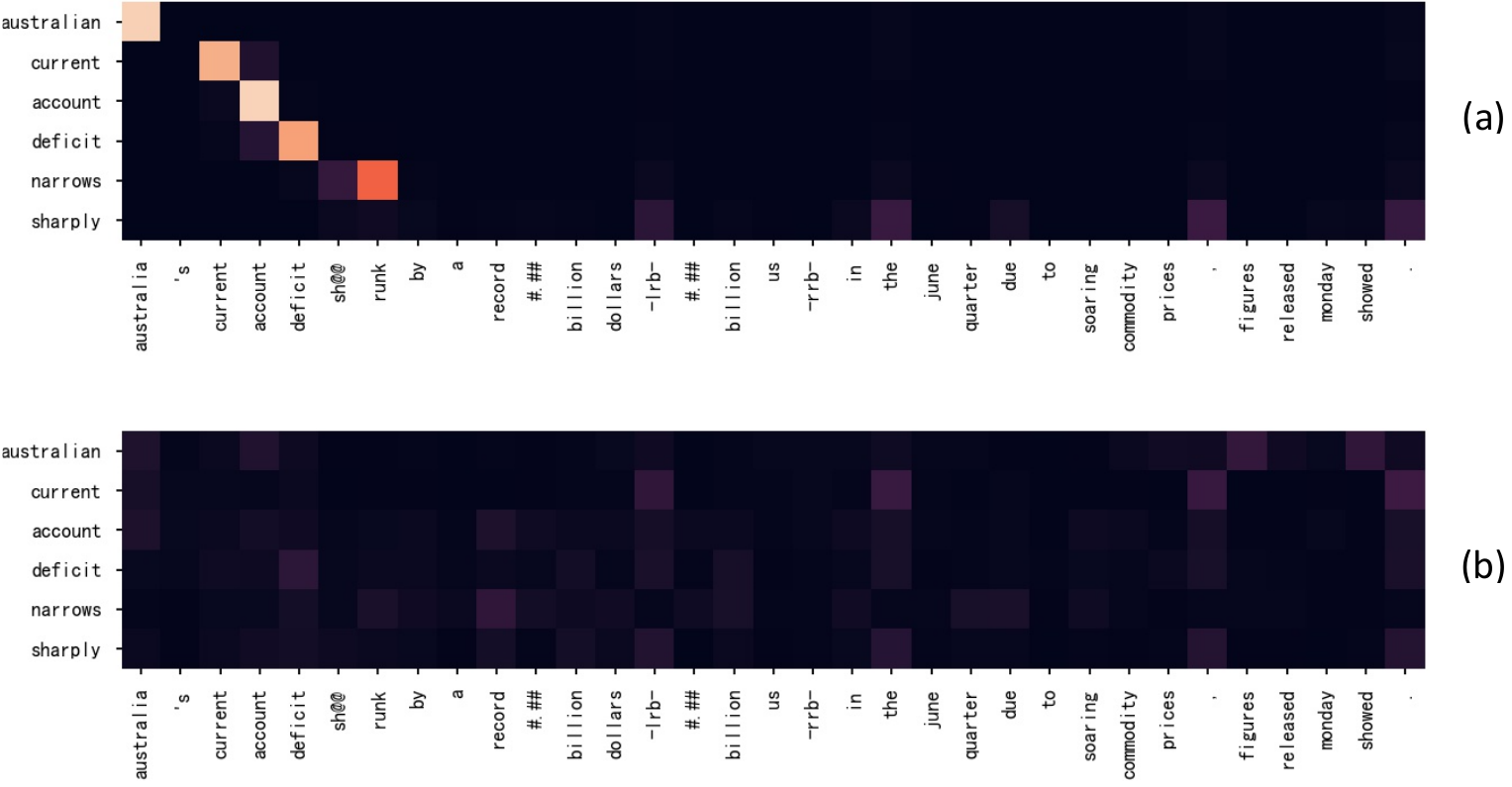}
\caption{Heatmaps of two sampled heads from the conventional encoder-decoder attention. (a) is of the fifth head of the third layer, and (b) is of the fifth head of the first layer.} 
\label{fig:heads}
\end{figure}

As illustrated in Figure \ref{fig:overall}, the opponent attention is derived from the conventional encoder-decoder attention. Since the multi-head attention is employed in Transformer, there are N$\times h$ heads in total in the conventional encoder-decoder attention, where N denotes the number of layers, $h$ denotes the number of heads in each layer. These heads exhibit diverse attention behaviors, posing a challenge of determining which head to derive the opponent attention, so that it attends to irrelevant or less relevant parts.

Figure \ref{fig:heads} illustrates the attention weights of two sampled heads. The attention weights in (a) well reflect the word level relevant relation between the source sentence and the target summary, while attention weights in (b) do not. We find that such behavior characteristic of each head is fixed. For example, head (a) always exhibits the relevant relation across different sentences and different runs. Based on depicting heatmaps of all heads for a few sentences, we choose the head that corresponds well to the relevant relation between source and target to derive the opponent attention \footnote{Given manual alignments between source and target of sampled sentence-summary pairs, we select the head that has the lowest alignment error rate (AER) of its attention weights.}.  

Specifically, let $\alpha_{c}$ denote the conventional encoder-decoder attention weights of the head which is used for deriving the opponent attention: 

\begin{equation}
\alpha_{c}={\rm softmax}(\frac{qk^{\rm T}}{\sqrt{d_k}})       
\end{equation}

\noindent where $q$ and $k$ are from the head same to that of $\alpha_c$. Let $\alpha_o$ denote the opponent attention weights. It is obtained through the opponent function applied on $\alpha_c$ followed by the softmax function: 

\begin{equation}
\alpha_o={\rm softmax}({\rm opponent}(\alpha_c))    \label{equ:alpha_o}
\end{equation}

The opponent function in Equation (\ref{equ:alpha_o}) performs a masking operation, which finds the maximum weight in $\alpha_c$, and replaces it with the negative infinity value, so that the softmax function outputs zero given the negative infinity value input. Then the maximum weight in $\alpha_c$ is set zero in $\alpha_o$ after the opponent and softmax functions. In this way, the most relevant part of the source sequence, which receives maximum attention in the conventional attention weights $\alpha_c$, is masked and neglected in $\alpha_o$. Instead, the remaining less relevant or irrelevant parts are extracted into $\alpha_o$ for the following contrastive training and decoding.

We also tried other methods to calculate the opponent attention weights, such as $\alpha_o = {\rm softmax}(1 - \alpha_c)$ ~\cite{Song2018Mask} \footnote{Song \textit{et al.} \shortcite{Song2018Mask} directly let $\alpha_o = 1 - \alpha_c$ in extracting background features for person re-identification in computer vision. We have to add softmax function since the attention weights must be normalized to one in sequence-to-sequence framework.} or $\alpha_o = {\rm softmax}(1 / \alpha_c)$, which aims to make $\alpha_o$ contrary to $\alpha_c$, but they underperform the masking opponent function on all benchmark datasets. So we present only the masking opponent in the following sections.

After $\alpha_o$ is obtained via Equation (\ref{equ:alpha_o}), the opponent attention is: ${\rm attention}_o = \alpha_o v$, where $v$ is from the head same to that of $q$ and $k$ in computing $\alpha_c$. 

Compared to the conventional attention ${\rm attention}_c$, which summarizes current relevant context, ${\rm attention}_o$ summarizes current irrelevant or less relevant context. They constitute a contrastive pair, and contribute together for the final summary word generation.

\subsubsection{Opponent Probability}

The opponent probability $P_o(y_i|y_1^{i-1},X)$ is computed by stacking several layers on top of ${\rm attention}_o$, and a softmin layer in the end as shown in the right part of Figure (\ref{fig:overall}). In particular,  

\begin{small}
\begin{eqnarray}
z_1 &=& {\rm LayerNorm}({\rm attention}_o)  \label{equ:att_o}  \\
z_2 &=& {\rm FeedForward}(z_1)   \\
z_3 &=& {\rm LayerNorm}(z_1 + z_2)   \\
P_o(y_i|y_1^{i-1},X) &=& {\rm softmin}(W z_3) \label{equ:p_o}
\end{eqnarray}
\end{small}

\noindent where $W$ is the matrix of the linear projection sublayer. 

${\rm attention}_o$ contributes to $P_o$ via Equation (\ref{equ:att_o}-\ref{equ:p_o}) step by step. The LayerNorm and FeedForward layers with residual connection is similar to the original Transformer, while a novel softmin function is introduced in the end to invert the contribution from ${\rm attention}_o$:

\begin{equation}
{\rm softmin}(v_i) = \frac{e^{(-v_i)}} { \sum^j e^{(-v_j)}}
\end{equation}

\noindent where $v=W z_3$, i.e., the input vector to the softmin function in Equation (\ref{equ:p_o}). Softmin normalizes $v$ so that scores of all words in the summary vocabulary sum to one. We can see that the bigger the $v_i$, the smaller the $P_{o,i}$ is.

Softmin functions contrarily to softmax. As a result, when we try to maximize $P_o(y_i = y | y_1^{i-1},X)$, where $y$ is the gold summary word, we effectively search for an appropriate ${\rm attention}_o$ to generate the lowest $v_g$, where $g$ is the index of $y$ in $v$. It means that the more irrelevant is ${\rm attention}_o$ to the summary, the lower the $v_g$ can be obtained, resulting in higher $P_o$.

\subsection{Training and Decoding}

During training, we jointly maximize the conventional probability $P_c$ and the opponent probability $P_o$:

\begin{equation}
\mathcal{J} = {\rm log}(P_c(y_i | y_1^{i-1},X) + \lambda {\rm log}(P_o(y_i | y_1^{i-1},X)  \label{equ:train}
\end{equation}

\noindent where $\lambda$ is the balanced weight. The conventional probability is computed as the original Transformer does, basing on sublayers of feed-forward, linear projection, and softmax stacked over the conventional attention as illustrated in the left part of Figure \ref{fig:overall}. The opponent probability is based on similar sublayers stacked over the opponent attention, but with softmin as the last sublayer as illustrated in the right part of Figure \ref{fig:overall}. 

Due to the contrary properties of softmax and softmin, jointly maximizing $P_c$ and $P_o$ actually maximizes the contribution from the conventional attention for summary word generation, while at the same time minimizes the contribution from the opponent attention\footnote{We also tried replacing softmin in Equation (\ref{equ:p_o}) with softmax, and correspondingly setting the training objective as maximizing $\mathcal{J} = {\rm log}(P_c) {\bf -} \lambda {\rm log}(P_o)$, but this method failed to train because $P_o$ becomes too small during training, and results in negative infinity value of ${\rm log}(P_o)$ that hampers the training. In comparison, softmin and the training objective of Equation (\ref{equ:train}) do not have such problem, enabling the effective training of the proposed network.}. In other words, the training objective is to let the relevant part attended by ${\rm attention}_c$ contribute more to the summarization, while let the irrelevant or less relevant parts attended by ${\rm attention}_o$ contribute less. 

During decoding, we aim to find maximum $\mathcal{J}$ of Equation (\ref{equ:train}) in the beam search process.

\section{Experiments}

We conduct experiments on abstractive sentence summarization benchmark datasets to demonstrate the effectiveness of the proposed contrastive attention mechanism. 

\subsection{Datasets}

In this paper, we evaluate our proposed method on three abstractive text summarization benchmark datasets. First, we use the annotated Gigaword corpus and preprocess it identically to Rush \textit{et al.} \shortcite{Rush2015Sum}, which results in around 3.8M training samples, 190K validation samples and 1951 test samples for evaluation. The source-summary pairs are formed through pairing the first sentence of each article with its headline. We use DUC-2004 as another English data set only for testing in our experiments. It contains 500 documents, each containing four human-generated reference summaries. The length of the summary is capped at 75 bytes. The last data set we used is a large corpus of Chinese short text summarization (LCSTS) ~\cite{Hu2015LCSTS}, which is collected from the Chinese microblogging website Sina Weibo. We follow the data split of the original paper, with 2.4M source-summary pairs from the first part of the corpus for training, 725 pairs from the last part with high annotation score for testing.

\begin{table*}[htb]
\small
  \centering
  \begin{tabular}{l | c c c | c c c }
    \multirow{2}{*}{System}
     & \multicolumn{3}{ c |}{Gigaword} & \multicolumn{3}{ c }{DUC2004} \\
    \cline{2-7}
    & R-1 & R-2 & R-L & R-1 & R-2 & R-L \\
    \hline\hline
    ABS~\cite{Rush2015Sum} & 29.55 & 11.32 & 26.42 & 26.55 & 7.06 & 22.05 \\
    ABS+~\cite{Rush2015Sum} & 29.76 & 11.88 & 26.96 & 28.18 & 8.49 & 23.81 \\
    RAS-Elman~\cite{Chopra2016RNN} & 33.78 & 15.97 & 31.15 & 28.97 & 8.26 & 24.06 \\
    words-lvt5k-1sent~\cite{Nallapati2016Abstractive} & 35.30 & 16.64 & 32.62 & 28.61 & 9.42 & 25.24 \\
    SEASS$_{\rm beam}$~\cite{Zhou2017Selective} & 36.15 & 17.54 & 33.63 & 29.21 & 9.56 & 25.51 \\
    RNN$_{\rm MRT}$~\cite{Ayana2016NHG} & 36.54 & 16.59 & 33.44 & 30.41 & 10.87 & 26.79 \\
    Actor-Critic ~\cite{Li2018Actor} & 36.05 & 17.35 & 33.49 & 29.41& 9.84 & 25.85 \\
    StructuredLoss~\cite{Edunov2018Classical} & 36.70 & 17.88 & 34.29 & - & - & - \\
    DRGD~\cite{Li2017Deep} & 36.27 & 17.57 & 33.62 & 31.79 & 10.75 & 27.48 \\
    ConvS2S~\cite{Gehring2017Convolutional}  & 35.88 & 17.48 & 33.29 & 30.44 & 10.84 & 26.90 \\
    ConvS2S$_{\rm ReinforceTopic}$~\cite{Wang2018CNN} & 36.92 & 18.29 & 34.58 & 31.15 & 10.85 & \textbf{27.68} \\
    FactAware~\cite{Cao2018Faithful} & 37.27 & 17.65 & 34.24 & - & - & - \\
    \hline
    Transformer & 37.87 & 18.69 & 35.22 & 31.38 & 10.89 & 27.18 \\
    Transformer+ContrastiveAttention & \textbf{38.72} & {\bf 19.09} & \textbf{35.82}  & \textbf{32.22} & \textbf{11.04} & 27.59 \\
  \end{tabular}
  \caption{ROUGE scores on the English evaluation sets of both Gigaword and DUC2004. On Gigaword, the full-length F-1 based ROUGE scores are reported. On DUC2004, the recall based ROUGE scores are reported. ``-" denotes no score is available in that work. }\label{tbl:exp_en}
\end{table*}

\subsection{Experimental Setup}

We employ Transformer as our basis architecture\footnote{https://github.com/pytorch/fairseq}. Six layers are stacked in both the encoder and decoder, and the dimensions of the embedding vectors and all hidden vectors are set 512. The inner layer of the feed-forward sublayer has the dimensionality of 2048. We set eight heads in the multi-head attention. The source embedding, the target embedding and the linear sublayer are shared in our experiments. Byte-pair encoding is employed in the English experiment with a shared source-target vocabulary of about 32k tokens ~\cite{Sennrich2015Neural}.

Regarding the contrastive attention mechanism, the opponent attention is derived from the head whose attention is most synchronous to word alignments of the source-summary pair. In our experiments, we select the fifth head of the third layer for deriving the opponent attention in the English experiments, and select the second head of the third layer in the Chinese experiments. All dimensions in the contrastive architecture are set 64. The $\lambda$ in Equation (\ref{equ:train}) is tuned on the development set in each experiment.

During training, we use the Adam optimizer with $\beta$1 = 0.9, $\beta$2 = 0.98, $\varepsilon$= 10$^{-9}$. The initial learning rate is 0.0005. The inverse square root schedule is applied for initial warm up and annealing ~\cite{Vaswani2017Attention}. During training, we use a dropout rate of 0.3 on all datasets.

During evaluation, we employ ROUGE ~\cite{Lin2004ROUGE} as our evaluation metric. Since standard Rouge package is used to evaluate the English summarization systems, we also follow the method of Hu \textit{et al.} \shortcite{Hu2015LCSTS} to map Chinese words into numerical IDs in order to evaluate the performance on the Chinese data set.

\subsection{Results}

\subsubsection{English Results}

The experimental results on the English evaluation sets are listed in Table \ref{tbl:exp_en}. We report the full-length F-1 scores of ROUGE-1 (R-1), ROUGE2 (R-2), and ROUGE-L (R-L) on the evaluation set of the annotated Gigaword, while report the recall-based scores of the R-1, R-2, and R-L on the evaluation set of DUC2004 to follow the setting of the previous works. 

The results of our works are shown at the bottom of Table \ref{tbl:exp_en}. The performances of the related works are reported in the upper part of Table \ref{tbl:exp_en} for comparison. ABS and ABS+ are the pioneer works of using neural models for abstractive text summarization. RAS-Elman extends ABS/ABS+ with attentive CNN encoder. words-lvt5k-1sent uses large vocabulary and linguistic features such as POS and NER tags. RNN$_{\rm MRT}$, Actor-Critic, StructuredLoss are sequence-level training methods to overcome the problem of the usual teacher-forcing methods. DRGD uses recurrent latent random model to improve summarization quality. FactAware generates summary words conditioned on both the source text and the fact descriptions extracted from OpenIE or dependencies. Besides the above RNN-based related works, CNN-based architectures of ConvS2S and ConvS2S$_{\rm ReinforceTopic}$ are included for comparison. 

Table \ref{tbl:exp_en} shows that we build a strong baseline using Transformer alone which obtains the state-of-the-art performance on Gigaword evaluation set, and obtains comparable performance to the state-of-the-art on DUC2004. When we introduce the contrastive attention mechanism into Transformer, it significantly improves the performance of Transformer, and greatly advances the state-of-the-art on both Gigaword evaluation set and DUC2004, as shown in the row of ``Transformer+Contrastive Attention".

\begin{table}[htb]
\centering
  \small
  \begin{tabular}{l | c c c }
  System & R-1 & R-2 & R-L  \\
  \hline\hline
  RNN context ~\cite{Hu2015LCSTS} & 29.90 & 17.40 & 27.20 \\
  CopyNet ~\cite{Gu2016Incorporating} & 34.40 & 21.60 & 31.30 \\
  RNN$_{\rm MRT}$~\cite{Ayana2016NHG} & 38.20 & 25.20 & 35.40 \\
  RNN$_{\rm distraction}$~\cite{Chen2016Distraction} & 35.20 & 22.60 & 32.50 \\
  DRGD ~\cite{Li2017Deep} & 36.99 & 24.15 & 34.21 \\
  Actor-Critic ~\cite{Li2018Actor} & 37.51 & 24.68 & 35.02 \\
  Global ~\cite{Lin2018Global} & 39.40 & 26.90 & 36.50 \\
  \hline
  Transformer & 41.93 & 28.28 & 38.32  \\
  Transformer+ContrastiveAttention  & \textbf{44.35} & \textbf{30.65} & \textbf{40.58}  \\
  \end{tabular}
  \caption{The full-length F-1 based ROUGE scores on the Chinese evaluation set of LCSTS. }\label{tbl:exp_cn}
\end{table}

\subsubsection{Chinese Results}

Table \ref{tbl:exp_cn} presents the evaluation results on LCSTS. The upper rows list the performances of the related works, the bottom rows list the performances of our Transformer baseline and the integration of the contrastive attention mechanism into Transformer. We only take character sequences as source-summary pairs and evaluate the performance based on reference characters for strict comparison to the related works.

Table \ref{tbl:exp_cn} shows that Transformer also sets a strong baseline on LCSTS that surpasses the performances of the previous works. When Transformer is equipped with our proposed contrastive attention mechanism, the performance is significantly improved and drastically advances the state-of-the-art on LCSTS.

\section{Analysis and Discussion}

\subsection{Effect of the Contrastive Attention Mechanism on Attentions}

\begin{figure}[htbp]
\flushleft
\includegraphics[height=8cm,width=8cm]{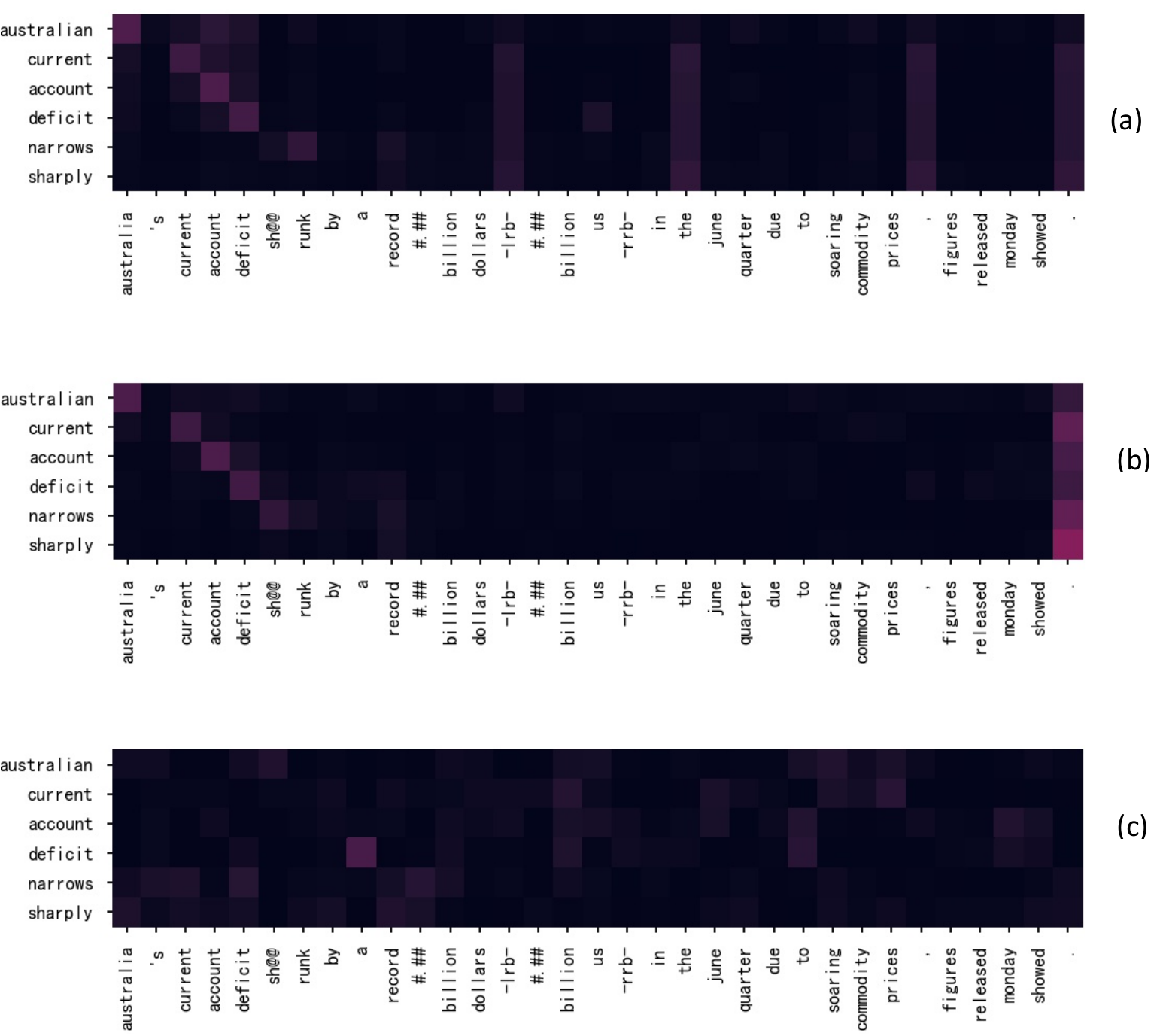}
\caption{The attention weight changes by using the contrastive attention mechanism. (a) is the average attention weights of the third layer of the baseline Transformer, (b) is that of ``Transformer+ContrastiveAttention", and (c) is the opponent attention derived from the fifth head of the third layer. }
\label{fig:att_results}
\end{figure}

Figure \ref{fig:att_results} shows the attention weights before and after using the contrastive attention mechanism. We depict the averaged attention weights of all heads in one layer in Figure \ref{fig:att_results}a and \ref{fig:att_results}b to study how it contributes to the conventional probability computation, and depict the opponent attention weights in Figure \ref{fig:att_results}c to study its contribution to the opponent probability. Since we select the fifth head of the third layer to derive the opponent attention in English experiment, the studies are carried out on the third layer.

Figure \ref{fig:att_results}a is from the baseline Transformer, Figure \ref{fig:att_results}b is from ``Transformer + ContrastiveAttention". We can see that ``Transformer + ContrastiveAttention" is more focused on the source parts that are most relevant to the summary than the baseline Transformer, which scatters attention weights on summary word neighbors or even functional words such as ``-lrb-" and ``the".  ``Transformer + ContrastiveAttention" cancels such scattered attentions by using the contrastive attention mechanism.

Figure \ref{fig:att_results}c depicts the opponent attention weights. They are optimized during training to generate the lowest score which is fed into softmin to get the highest opponent probability $P_o$. The more irrelevant to the summary word the opponent is, the lower the score can be obtained, thus resulting in higher $P_o$. Figure \ref{fig:att_results}c shows that the attentions are formed over irrelevant parts with varied weights as the result of maximizing $P_o$ during training.

\begin{table*}[htb]
\small
  \centering
  \begin{tabular}{l | c c c | c c c }
    \multirow{2}{*}{System}
     & \multicolumn{3}{ c |}{Gigaword} & \multicolumn{3}{ c }{DUC2004} \\
    \cline{2-7}
    & R-1 & R-2 & R-L & R-1 & R-2 & R-L \\
    \hline\hline
    mask maximum weight &  38.72 & 19.09 & 35.82 & 32.22 & 11.04 & 27.59  \\
    mask top-2 weights & 38.17&19.15 &35.51 &31.87  & 10.94 &27.41   \\
    mask top-3 weights &38.36 &19.11  &35.56 &31.67 &10.37 &27.31   \\
    dynamically mask &38.12 &18.92 & 35.28 & 31.37 &10.32 &27.11  \\
    \hline
    synchronous head  & 38.72 & 19.09 & 35.82  & 32.22 & 11.04 & 27.59 \\
    non-synchronous head  & 37.85 & 18.59  & 35.16  &  31.73 & 10.74  & 27.35  \\
    averaged head  & 38.43 & 19.10 & 35.53 & 31.82 & 10.98 & 27.43 \\
    \hline
    Transformer baseline & 37.87 & 18.69 & 35.22 & 31.38 & 10.89 & 27.18  \\
  \end{tabular}
  \caption{Results of explorations on the opponent attention derivation. The upper part presents the influence of masking more attention weights for deriving the opponent attention. The middle part presents the results of selecting different head for the opponent attention derivation. The bottom row presents the result of Transformer.}\label{tbl:explore_derivation}
\end{table*}

\begin{table}[htb]
\centering
  \small
  \begin{tabular}{l | c c c }
  \hline
  Gigaword & R-1 & R-2 & R-L  \\
  \hline
  Transformer & 37.87 & 18.69 & 35.22 \\
  Transformer+ContrastiveAtt-$P_o$ & 37.92 & 18.88 & 35.21  \\
  Transformer+ContrastiveAtt & 38.72 & 19.09 & 35.82 \\
  \hline\hline
  DUC2004  & R-1 & R-2 & R-L  \\
  \hline
  Transformer &31.38 & 10.89 & 27.18  \\
  Transformer+ContrastiveAtt-$P_o$ & 31.21 & 10.70 & 26.85 \\
  Transformer+ContrastiveAtt & 32.22 & 11.04 & 27.59  \\
  \hline
  \end{tabular}
  \caption{The effect of dropping $P_o$ (denoted by -$P_o$) from Transformer+ContrastiveAtt during decoding. }\label{tbl:exp_po}
\end{table}

\begin{table*}[htb]
\centering
  \small
  \begin{tabular}{p{2.1\columnwidth}}
  \textbf{Src}:press freedom in algeria remains at risk despite the release on wednesday of prominent newspaper editor mohamed UNK after a two-year prison sentence , human rights organizations said .   \\
  \textbf{Ref}:algerian press freedom at risk despite editor 's release UNK picture \\
  \textbf{Transformer}:press freedom remains at risk in algeria {\color{red} rights groups say} \\
  \textbf{Transformer+ContrastiveAtt}:press freedom remains at risk {\color{blue} despite release of algerian editor}  \\
  \hline
  \textbf{Src}:denmark 's poul-erik hoyer completed his hat-trick of men 's singles badminton titles at the european championships , winning the final here on saturday  \\
  \textbf{Ref}:hoyer wins singles title \\
  \textbf{Transformer}:hoyer {\color{red} completes hat-trick} \\
  \textbf{Transformer+ContrastiveAtt}:hoyer {\color{blue} wins men 's singles title}  \\
  \hline
  \textbf{Src}:french bank credit agricole launched on tuesday a public cash offer to buy the \#\# percent of emporiki bank it does not already own , in a bid valuing the greek group at \#.\# billion euros ( \#.\# billion dollars ) .\\
  \textbf{Ref}:credit agricole announces \#.\#-billion-euro bid for greek bank emporiki  \\
  \textbf{Transformer}:credit agricole {\color{red} launches public cash offer} for greek bank  \\
  \textbf{Transformer+ContrastiveAtt}:french bank credit agricole {\color{blue} bids \#.\# billion euros} for greek bank  \\
 \end{tabular}
 \caption{Example summaries generated by the baseline Transformer and Transformer+ContrastiveAtt. }\label{tbl:examples}
\end{table*}

\subsection{Effect of the Opponent Probability in Decoding}

We study the contribution of the opponent probability $P_o$ by dropping it during decoding to see if it hurts the performance. Table \ref{tbl:exp_po} shows that dropping $P_o$ significantly harms the performance of ``Transformer + ContrastiveAtt". The performance difference between the model dropping $P_o$ and the baseline Transformer is marginal, indicating that adding the opponent probability $P_o$ is key for achieving the performance improvement.  

\subsection{Explorations on Deriving the Opponent Attention}

\vspace{6 pt}
\noindent  \textbf{Masking More Attention Weights for Deriving the Opponent Attention}

\noindent In Section \ref{sec:opponent_att}, we mask the most salient word that has the maximum weight of $\alpha_c$ to derive the opponent attention. In this subsection, we experimented with masking more weights of $\alpha_c$ by two ways: 1) masking top $k$ weights, 2) dynamically masking. In the dynamically masking method, we order the weights from big to small at first, then go on masking two neighbors until the ratio between them is over a threshold. The threshold is 1.02 based on training and tuning on the development set.

The upper rows of Table \ref{tbl:explore_derivation} presents the performance comparison between masking maximum weight and masking more weights. It shows that masking maximum weight performs better, indicating that masking the most salient weight leaves more irrelevant or less relevant words to compute the opponent probability $P_o$, which is more reliable than that computed from less remaining words after masking more weights.

\vspace{6 pt}
\noindent  \textbf{Selecting Non-synchronous Head or Averaged Head for Deriving the Opponent Attention}

\noindent As explained in Section \ref{sec:opponent_att}, the opponent attention is derived from the head that is most synchronous to the word alignments between source sentence and summary. We denote it ``synchronous head". We also explored deriving the opponent attention from the fifth head of the first layer, which is non-synchronous to the word alignments as illustrated in Figure \ref{fig:heads}b. Its result is presented in the ``non-synchronous head" row. In addition, the attention weights averaged on all heads of the third layer are used to derive the opponent attention. We denote it ``averaged head". 

As shown in the middle part of Table \ref{tbl:explore_derivation}, both ``non-synchronous head" and ``averaged head" underperform ``synchronous head". ``non-synchronous head" performs worst, and even worse than the Transformer baseline on Gigaword. This indicates that it is better to compose the opponent attention from irrelevant parts that can be easily located in the synchronous head. ``averaged head" performs slightly worse than ``synchronous head", and is also slower due to the involved all heads.

\subsection{Qualitative Study}

Table \ref{tbl:examples} shows the qualitative results. The highlights in the baseline Transformer manifest the incorrect areas extracted by the baseline system. In contrast, the highlights in Transformer+ContrastiveAtt show that correct contents are extracted since the contrastive system distinguish relevant parts from irrelevant parts on the source side and made attending to correct areas more easily.

\section{Conclusion}

We proposed a contrastive attention mechanism for abstractive sentence summarization, using both the conventional attention that attends to the relevant parts of the source sentence, and a novel opponent attention that attends to irrelevant or less relevant parts for the summary word generation. Both categories of the attention constitute a contrastive pair, and we encourage contribution from the conventional attention and penalize contribution from the opponent attention through joint training. Using Transformer as a strong baseline, experiments on three benchmark data sets show that the proposed contrastive attention mechanism significantly improves the performance, advancing the state-of-the-art performance for the task.

\section*{Acknowledgments}

The authors would like to thank the anonymous reviewers for the helpful comments. This work was supported by National Key R\&D Program of China (Grant No. 2016YFE0132100), National Natural Science Foundation of China (Grant No. 61525205, 61673289).

\bibliography{emnlp-ijcnlp-2019_v2}
\bibliographystyle{acl_natbib}

\end{document}